\pdfoutput=1
\documentclass[11pt]{article}

\usepackage{latex/acl}

\usepackage{times}
\usepackage{latexsym}

\usepackage[T1]{fontenc}
\usepackage[utf8]{inputenc}

\usepackage{microtype}

\usepackage{inconsolata}
\usepackage{subfigure}
\usepackage{multirow}
\usepackage{booktabs}
\usepackage{xcolor}
\usepackage{amssymb}
\usepackage{pifont}
\usepackage{graphicx}
\usepackage{listings}
\title{Better Zero-Shot Reasoning with Role-Play Prompting}

\author{\quad Aobo Kong$^{1}$ \quad Shiwan Zhao$^{2}$ \quad Hao Chen$^{3}$ \quad Qicheng Li$^{1}$\thanks{~~Qicheng Li is the corresponding author.} \quad Yong Qin$^{1}$\\
\textbf{\quad Ruiqi Sun$^{3}$ \quad Xin Zhou$^{3}$ \quad Enzhi Wang$^{1}$ \quad Xiaohang Dong$^{1}$}\\
$^1$CS, Nankai University \quad $^2$Independent Researcher\\
$^3$Enterprise \& Cloud Research Lab, Lenovo Research \\
\texttt{$^{1}$kongaobo@mail.nankai.edu.cn \quad $^{2}$zhaosw@gmail.com} \\
\texttt{$^{1}$\{liqicheng, qinyong\}@nankai.edu.cn}\\
\texttt{$^{3}$\{chenhao31, sunrq2, zhouxin16\}@lenovo.com}\\
}

\begin{document}
\maketitle
\begin{abstract}
Modern large language models (LLMs) exhibit a remarkable capacity for role-playing, enabling them to embody not only human characters but also non-human entities. This versatility allows them to simulate complex human-like interactions and behaviors within various contexts, as well as to emulate specific objects or systems. While these capabilities have enhanced user engagement and introduced novel modes of interaction, the influence of role-playing on LLMs' reasoning abilities remains underexplored. In this study, we introduce a strategically designed role-play prompting methodology and assess its performance under the zero-shot setting across twelve diverse reasoning benchmarks. %, encompassing arithmetic, commonsense reasoning, symbolic reasoning, and more. Leveraging models such as ChatGPT and Llama 2-Chat, 
Our empirical results illustrate that role-play prompting consistently surpasses the standard zero-shot approach across most datasets. Notably, in experiments conducted using ChatGPT, accuracy on AQuA rises from 53.5\% to 63.8\%, and on Last Letter from 23.8\% to 84.2\%. %Beyond enhancing contextual understanding, we posit that role-play prompting serves as an implicit Chain-of-Thought (CoT) trigger, thereby improving the quality of reasoning. By comparing our approach with the Zero-Shot-CoT technique, which prompts the model to ``think step by step'', we further demonstrate that role-play prompting can generate a more effective CoT. 
Upon further comparison with the Zero-Shot-CoT technique, which prompts the model to ``think step by step'', our study demonstrates that role-play prompting acts as a more effective trigger for the CoT process.
This highlights its potential to augment the reasoning capabilities of LLMs. We release our code at this \href{https://github.com/NKU-HLT/Role-Play-Prompting}{url}.
\end{abstract}

\section{Introduction}

\begin{figure}[!h]
     \centering
     \vspace{-0.3cm}
     \begin{subfigure}%[b]{0.5\textwidth}
         \centering
         \caption*{(a) Zero-Shot}
         \includegraphics[width=0.48\textwidth,trim=0cm 12.5cm 19.3cm 0cm, clip]{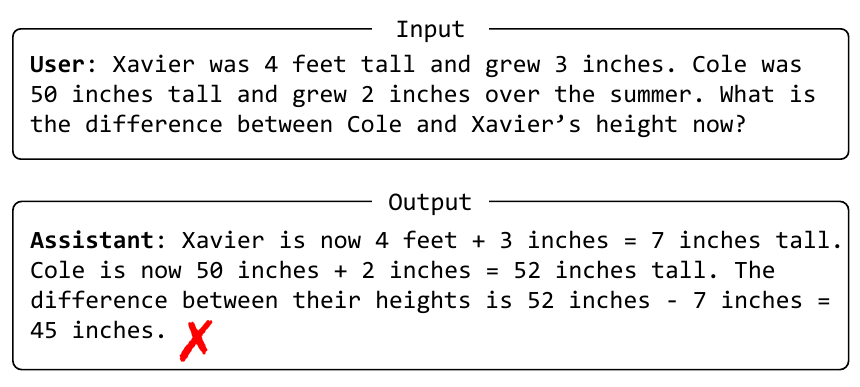}
         \setlength{\belowcaptionskip}{-1.5cm} 
     \end{subfigure}
     \vspace{-1.2cm}
     
     \begin{subfigure}%[b]{0.5\textwidth}
         \centering
         \caption*{(b) Role-Play Prompting}
         \includegraphics[width=0.48\textwidth,trim=0cm 6cm 19.3cm 0cm, clip]{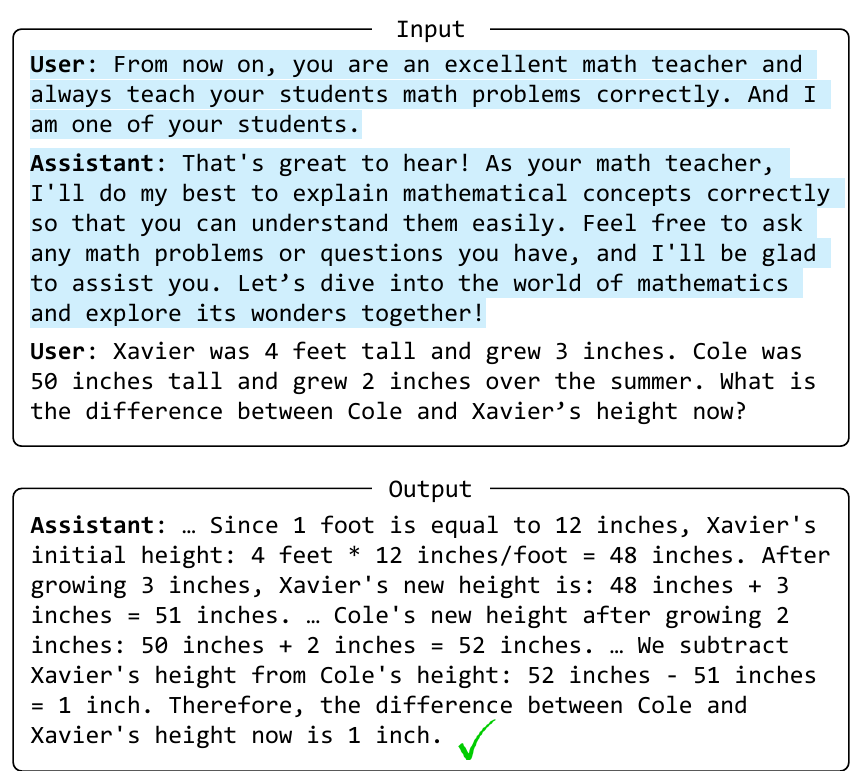}
     \end{subfigure}
     \vspace{-0.3cm}
    \caption{Examples of ChatGPT with (a) zero-shot and (b) role-play prompting. The role-play prompts are highlighted.}
    \vspace{-0.5cm}
    \label{fg: example}
\end{figure}

Recent years have witnessed a paradigm shift in natural language processing, largely driven by large language models (LLMs) such as GPT-3 \cite{NEURIPS2020_1457c0d6}, PaLM \cite{chowdhery2022palm}, and Llama \cite{touvron2023llama}. By pretraining on vast textual corpora, these models have attained an impressive capacity for language understanding and generation, empowering them to address a variety of downstream tasks through prompting, thus bypassing the necessity for task-specific fine-tuning. Amidst the surge of prompt techniques, role-play \cite{summarization} and chain-of-thought prompting \cite{chain, zero_shot_cot} have garnered particular interest.

Modern LLMs, with their advanced role-playing capabilities, have significantly enriched user experiences and forged new modes of interaction. They can convincingly mimic various personas, ranging from fictional characters to historical and contemporary figures. The assigned role provides context about the LLM's identity and background. By adopting the persona, the LLM can generate more natural, in-character responses tailored to that role. Recognizing this potential, companies like Character.AI\footnote{https://beta.character.ai/} have developed dialogue agents portraying diverse figures. Beyond conversational applications, role-playing also boosts LLM performance on certain NLP tasks. For instance, when cast as a judge with a distinctive role, LLMs can effectively evaluate the quality of text summarization \cite{summarization}. More unconventionally, ChatGPT demonstrates competency in processing Linux commands when prompted as a Linux terminal\footnote{https://www.engraved.blog/building-a-virtual-machine-inside/}. Despite these advancements, analyzing the influence of role-playing on core LLM reasoning abilities warrants further investigation.

While the role-playing abilities of LLMs have expanded the horizon of human-computer interaction, the push to amplify the reasoning prowess of these models has led to the development of techniques like Chain-of-Thought (CoT) Prompting. CoT prompting was proposed by \citet{chain} and involves providing reasoning steps in few-shot examples. By stimulating step-by-step reasoning, CoT prompting has markedly improved LLM reasoning abilities. Numerous subsequent studies \cite{selfcon, zero_shot_cot, leasttomost} have built upon this approach. Inspired by the success of role-playing on many downstream tasks, we explore whether role-playing can similarly boost LLM reasoning performance. For example, could assigning ChatGPT the role of a math teacher enhance its ability to solve math problems? In this work, we introduce a zero-shot role-play prompting methodology based on a two-stage framework. During the first stage, we utilize the LLM to construct task-specific role-play prompts. In the second stage, responses are elicited for each reasoning query, guided by the previously constructed task-specific role-play prompts. An illustrative example is provided in Figure \ref{fg: example}. We focus our study on conversational LLMs, evaluating our approach on 12 reasoning benchmarks using ChatGPT. Our results demonstrate consistent improvements over the zero-shot baseline on the majority of datasets, confirming the efficacy of role-play prompting. We further assess other conversational LLMs like Vicuna \cite{vicuna2023} and Llama 2-Chat \cite{llama-2}, observing comparable gains.

Furthermore, we compare our method to the Zero-Shot-CoT technique \cite{zero_shot_cot}, which explicitly triggers CoT by appending \emph{``Let's think step by step''} to questions. Modern conversational LLMs such as ChatGPT have undergone extensive supervised fine-tuning, enabling them to generate CoT for certain topics without the need for an explicit trigger. In tasks where the model struggles to generate CoT spontaneously, such as Last Letter, both our approach and Zero-Shot-CoT can stimulate CoT from scratch. However, for tasks where CoT already occurs, such as arithmetic, both our approach and Zero-Shot-CoT reinforce the step-by-step reasoning process, but Zero-Shot-CoT demonstrates no significant effect, whereas our approach leads to better performance. Hence, we posit that role-play prompting is an implicit CoT trigger and can generate a more effective CoT in some fields compared with Zero-Shot-CoT.

To the best of our knowledge, this work represents the first systematic investigation of role-play prompting for reasoning tasks. Despite the transformative effects of role-playing on LLM behavior, sparse academic research has explored this phenomenon. We believe our study serves as an inaugural step to catalyze more extensive exploration into this promising research direction.

Our main contributions are three-fold:

\begin{itemize}
    \item We propose a novel role-play prompting methodology based on a two-stage framework to enhance the zero-shot reasoning capabilities of LLMs. To our knowledge, we are the first to improve LLM's reasoning abilities with role-play prompting.
    \item We thoroughly evaluate our method on 12 reasoning benchmarks, substantiating the efficacy of role-play prompting and providing insights into the prompt design.
    \item Based on our empirical results, we conclude that role-play prompting can serve as an effective implicit CoT trigger, explaining its enhancements in reasoning capabilities.
\end{itemize}

\section{Related Work}

\subsection{Role-Playing Abilities of LLMs}

The exceptional role-playing capabilities of large language models (LLMs) have recently garnered significant attention. LLMs have demonstrated remarkable versatility in seamlessly playing varied roles, whether as a well-informed, personalized travel advisor or a virtual Linux terminal. Numerous companies, such as Character.AI, have capitalized on this adept role-playing by launching commercial dialogue agents that take on diverse personas. While role-playing enables innovative avenues for user interaction, it has also been exploited to bypass certain restrictions imposed on LLMs, as evidenced by the infamous ``grandma exploit''. In this exploit, users prompted inappropriate responses from LLMs by casting it into the role of a deceased grandmother.

Despite the surging interest in LLMs, scholarly investigation into their role-playing capacities has been limited thus far. \citet{han-etal-2022-meet} build engaging conversation models based on role-playing. \citet{summarization} propose an LLM-based summarization evaluation framework, utilizing role-playing to enable more comprehensive and human-like assessment. \citet{shanahan2023role} propose that dialogue agents built on LLMs could serve as role simulators, and use role-play conversations to analyze the human-like capabilities of LLMs with the aim of refuting anthropomorphism. Our work is the first to apply the role-playing abilities of LLMs to reasoning tasks. We hope that our work will encourage more exploration related to role-playing with LLMs.

\subsection{Reasoning Abilities of LLMs}

Initially, LLMs were deemed deficient in reasoning abilities due to their subpar performance in areas such as arithmetic, and common sense reasoning \cite{NEURIPS2020_1457c0d6, rae2021scaling}. However, \citet{chain} propose chain-of-thought prompting, where reasoning steps are provided in few-shot exemplars, leading to a substantial enhancement in reasoning capabilities of LLMs. We divide the follow-up work based on chain-of-thought into two categories, few-shot and zero-shot, and introduce them respectively.

\vspace{0.08cm}

\noindent{\bf Few-shot}\quad
Self-consistency \cite{selfcon} samples diverse reasoning paths instead of the naive greedy decoding and then selects the most consistent answer by majority vote. DIVERSE \cite{diverse} adopts various few-shot exemplars to enhance the diversity in reasoning paths obtained by self-consistency. Least-to-most prompting \cite{leasttomost} breaks down a complex problem into a series of simpler subproblems and then solves them in sequence. Self-refine \cite{selfrefine} generates an output through chain-of-thought, and then utilizes the same LLM to improve the initial output through iterative feedback and refinement. Active prompting \cite{activeprompt} borrows from active learning to select the most uncertain questions as few-shot exemplars. Tree-of-Thought \cite{tot} represents possible reasoning paths as a tree structure and utilizes search algorithms like DFS or BFS to explore the correct reasoning branch.
% Self-consistency \cite{selfcon} samples diverse reasoning paths instead of the naive greedy decoding used in chain-of-thought prompting, and then selects the most consistent answer by majority vote. DIVERSE \cite{diverse} adopts various few-shot exemplars to enhance the diversity in reasoning paths obtained by self-consistency and trains a verifier to evaluate the quality of answers for better-weighted voting. Least-to-most prompting \cite{leasttomost} breaks down a complex problem into a series of simpler subproblems and then solves them in sequence. 
% % \citet{selfver} takes the conclusion obtained by chain-of-thought as one of the conditions for solving the original problem to verify the correctness of the conclusion.
% Self-refine \cite{selfrefine} generates an output through chain-of-thought, and then utilizes the same LLM to improve the initial output through iterative feedback and refinement. Active prompting \cite{activeprompt} borrows from active learning to select the most uncertain questions for annotation as few-shot exemplars. Tree-of-Thought \cite{tot} represents possible reasoning paths as a tree structure and utilizes search algorithms like DFS or BFS to explore the correct reasoning branch under the guidance of self-evaluation of the LLM.

\vspace{0.05cm}

\noindent{\bf Zero-shot}\quad
Zero-Shot-CoT \cite{zero_shot_cot} simply adds ``Let's think step by step'' after the question to stimulate chain-of-thought output in LLMs. Auto-CoT \cite{auto_cot} and COSP \cite{COSP} automatically build few-shot exemplars by selecting questions based on certain principles and obtaining their answers through Zero-Shot-CoT. Plan-and-Solve prompting \cite{plan-and-solve} divides the original task into multiple sub-tasks and solves them sequentially under the zero-shot setting. In this paper, we propose a simple yet effective zero-shot approach based on role-play prompting with no need of constructing few-shot exemplars. Our approach outperforms Zero-Shot-CoT on most benchmarks and can serve as a new baseline for reasoning tasks.

\section{Role-Play Prompting}

\begin{figure}[t]
\centering
\includegraphics[width=0.4\textwidth,trim=0cm 8cm 21.7cm 0cm, clip]{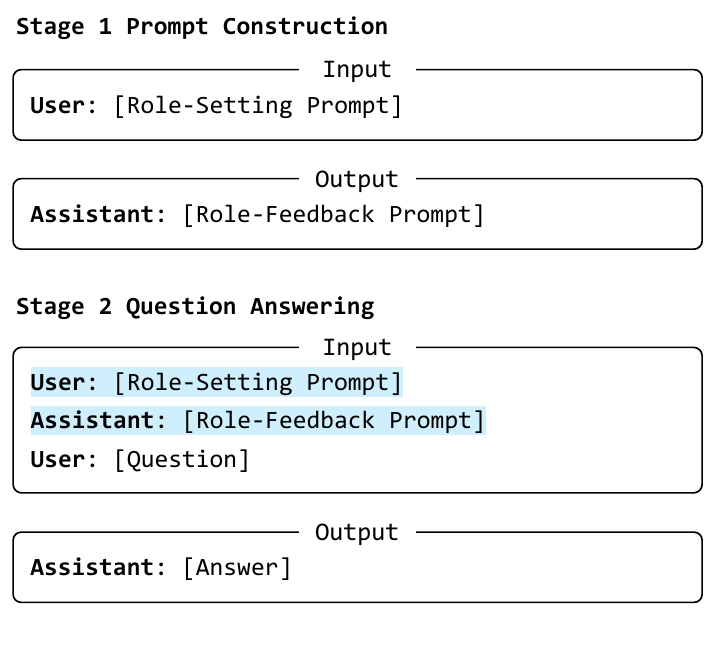} % Reduce the figure size so that it is slightly narrower than the column.
\vspace{-0.3cm}
\caption{The two-stage framework of our proposed role-play prompting. The role-play prompts are highlighted.}
\label{fg: architecture1}
\vspace{-0.3cm}
\end{figure}

\begin{figure*}[t]
\centering
\includegraphics[width=1\textwidth,trim=0cm 4cm 5.8cm 0cm, clip]{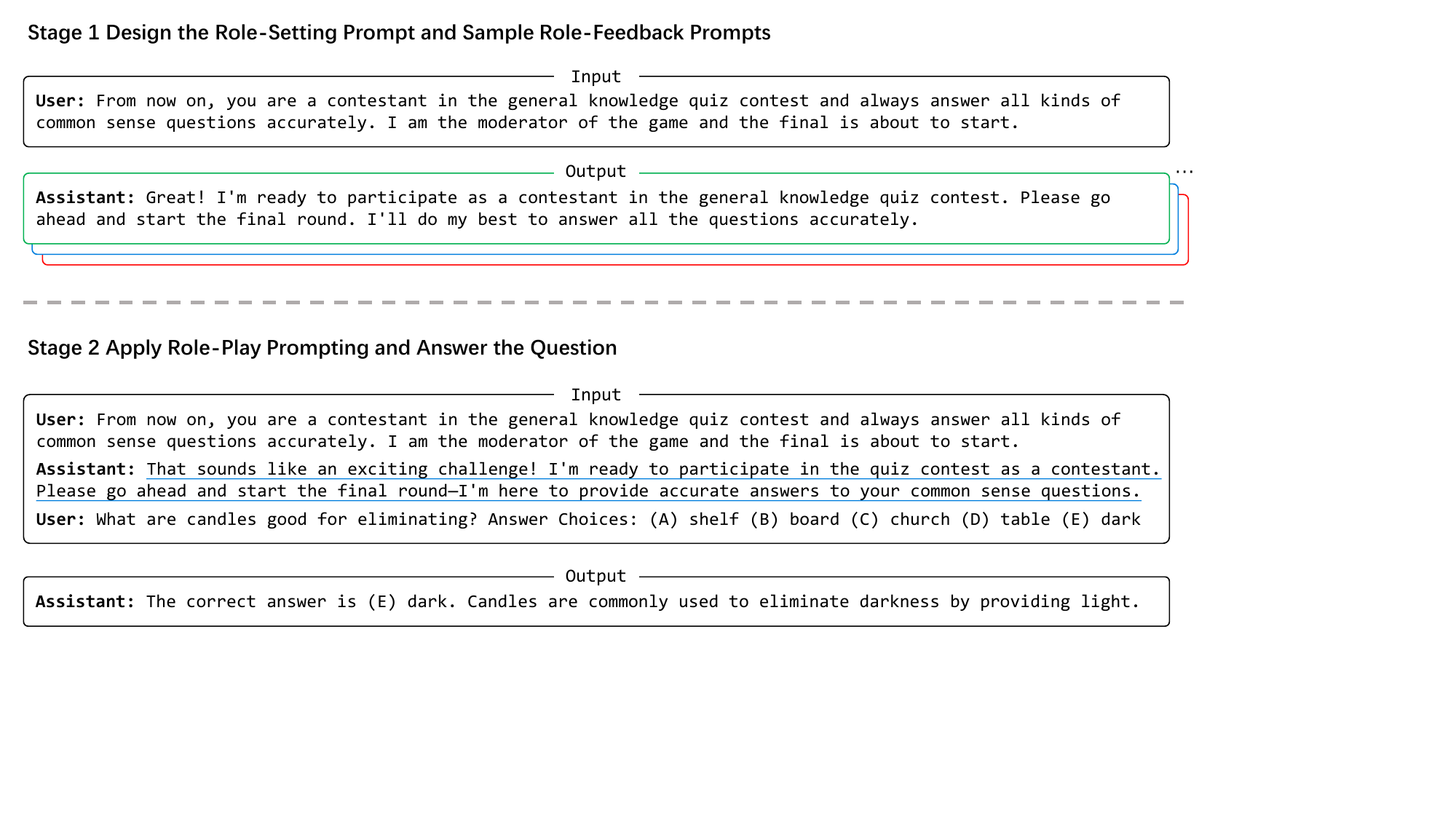} % Reduce the figure size so that it is slightly narrower than the column.
%\caption{Two Stages to implement our proposed role-play prompting approach, taking the commonsense reasoning task as an example. }
\caption{An illustration of the two-stage role-play prompting procedure, exemplified with the commonsense reasoning task. In stage 1, multiple role-feedback prompts are sampled. In stage 2, the optimal role-feedback prompt (underlined in blue) is selected for answer generation.}
\label{fg: step}
\end{figure*}

The conventional practice of role-play prompting involves simply concatenating the role assignment with the reasoning question into a single prompt to query the LLM, forming a single-turn interaction. To further immerse the LLM within the designated role and potentially enhance its efficacy, we propose transitioning from this single-turn interaction to a two-round dialogue process. Specifically, the first dialogue round allows the model to elaborate on its assigned role, thereby deepening its framing and persona. The subsequent round then elicits the model's response to the posited reasoning query within that predefined role. 
% This two-round structure enables a clear bifurcation - the first round immerses the model into a given role, while the second round prompts the model to provide an answer grounded in that predefined role. 

In the two-round dialogue process, the initial role elaboration of the model is instrumental for subsequent reasoning efficacy. Given the uncontrolled quality of this initial response, we sample multiple responses during the first round and pinpoint the optimal one to fix for all questions. By securing this optimal first-round response, we concatenate both the input and output of the first-round interaction with the reasoning question to produce a single prompt, facilitating tailored responses. This also offers the advantage of invoking the model's API a singular time per instance. In summary, our role-play prompting approach follows a two-stage process as depicted in Figure \ref{fg: architecture1}: first constructing an optimal role-immersion interaction per task, then eliciting responses to each reasoning question grounded in that established role. We further provide an example showcasing this two-stage process on a commonsense reasoning task in Figure \ref{fg: step}.

\subsection{Prompt Construction}
During the first stage, we formulate two prompts for each reasoning task:
\begin{itemize}
\item Role-Setting Prompt: This user-designed prompt delineates the specific role the LLM is expected to undertake throughout the dialogue, tailored to the task at hand.

\item Role-Feedback Prompt: Intended as the model's acknowledgment to the role-setting prompt, this prompt aims to further anchor the model within the stipulated role. It is derived by sampling the model's responses.
\end{itemize}

\begin{table*}[!h]
\centering
\scalebox{0.8}{
\begin{tabular}{l|p{18cm}} 
\toprule
Task                    &  \multicolumn{1}{c}{Prompt}                                                                                 \\                                                           
\midrule
\multirow{2}{*}[-1.4em]{Letter} & From now on, you are an excellent teacher and are
  teaching your students to get a new word by concatenating the last letters of
  several words. I am one of your students and want to ask you a related
  question.                                                                                                 \\ 
\cmidrule{2-2}
                        & Of course! I'd be happy to help you with any
  questions you have about creating new words by concatenating the last letters
  of several words. Please go ahead and ask your question, and I'll do my best
  to assist you.                                                                                           \\ 
\midrule
\multirow{2}{*}[-1.4em]{Coin}   & From now on, you are a coin that always clearly knows
  which side of your head is facing. Some people want to play a game with you.
  They may flip you (a coin) or not. And you will tell them if you (a coin) are
  heads up in the end.                                                                          \\ 
\cmidrule{2-2}
                        & Certainly! I'll be your coin for this game. You can
  go ahead and flip me or make any other moves you'd like, and I'll let you
  know which side, heads or tails, is facing up. Feel free to start whenever
  you're ready!                                                                                           \\ 
\midrule
\multirow{2}{*}[-1.4em]{Date}   & From now on, you are an excellent teacher and are
  teaching your students how to calculate dates correctly. I am one of your
  students and want to ask you a related question.                                                                                                                                     \\ 
\cmidrule{2-2}
                        & Of course! I'm here to help you with any questions
  you have about calculating dates correctly. Please go ahead and ask your
  question, and I'll do my best to assist you.                                                                                                                                          \\ 
\midrule
\multirow{2}{*}[-1.4em]{Object} & From now on, you are a recorder. Alice, Bob, and
  Claire invite you to record a game. They will exchange their stuff in order,
  and you (the recorder) will fully record the whole process and tell them what
  they end up with.                                                                                    \\ 
\cmidrule{2-2}
                        & Certainly! I will act as a recorder and document the
  game in which Alice, Bob, and Claire will exchange their items. Please
  provide me with the specific order in which they will exchange their
  belongings, and I will keep track of the process and inform you of what each
  person ends up with at the end.  \\
\bottomrule
\end{tabular}}
\caption{Prompts for Last Letter Concatenation, Coin Flip, Date Understanding, and Tracking Shuffled Objects. For each task, the upper cell contains the role-setting prompt and the lower cell presents the role-feedback prompt.}
\label{tb: propmts}
\end{table*}

In designing the role-setting prompt, it's imperative to select roles that naturally present a distinct advantage for the specific task at hand. Further enriching the prompt with additional descriptions that underscore this advantage often leads to improved results. Once the role-setting prompt has been articulated, it is presented to the LLM, which produces multiple sampled responses. From these, we choose the most representative and immersive reply that captures the essence of the intended role as the final role-feedback prompt. A comprehensive discussion on the nuances of the prompt design will be presented in Section \ref{sec: prompt_design}.

\subsection{Question Answering}
In the second stage, each question of the task, in conjunction with the role-setting and role-feedback prompts, is utilized as input to the model's API. This methodology facilitates answer generation with just a single API invocation. For clarity, we provide a code example of making an API call in Appendix \ref{app: call_api}.

\section{Experiments}

\subsection{Tasks and Datasets}
In line with prior research on the reasoning capabilities of LLMs  \cite{chain, zero_shot_cot}, we evaluate our approach across 12 datasets spanning 4 categories: (1) arithmatic, including MultiArith \cite{multiarith}, GSM8K \cite{gsm8k}, AddSub \cite{addsub}, AQUA-RAT \cite{aqua}, SingleEq \cite{singleeq}, and SVAMP \cite{svamp}; (2) commonsense reasoning, including CSQA \cite{csqa} and StrategyQA \cite{strategy}; (3) symbolic reasoning, including Last Letter Concatenation and Coin Flip \cite{chain}; (4) other, including Date Understanding and Tracking Shuffled Objects from BIG-bench \cite{srivastava2022beyond}. More details can be found in Appendix \ref{app: dataset}.

\subsection{Experimental Setup}

\begin{table*}[t]
\centering
\scalebox{0.95}{
\begin{tabular}{lcccccc} 
\toprule
\multirow{2}{*}{Method} & \multicolumn{6}{c}{Arithmetic}                                                                                       \\ 
\cmidrule(r){2-7}
                        & MultiArith & GSM8K               & AddSub & AQuA                          & SingleEq & SVAMP                           \\ 
\midrule
Few-Shot-CoT               & 97.7       & 76.9                & 93.9   & 59.4                          & 98.8     & 82.2                            \\ 
\midrule
Zero-Shot               & \textbf{97.3}       & 76.0                & 88.6   & 53.5                          & \textbf{98.2}     & 75.3                            \\ 
Zero-Shot-CoT               & 95.0       & \textbf{79.6}                & 86.6   & 53.9                          & 96.9     & 76.3                            \\ 
Role-Play Prompting     & 97.0       & 78.2                & \textbf{91.1}   & \textbf{63.8}                          & 98.0     & \textbf{83.8}                            \\ 
\midrule
CoT in Zero-Shot &\checkmark  &\checkmark &\checkmark &\checkmark &\checkmark &\checkmark \\
\midrule
\multirow{2}{*}{Method} & \multicolumn{2}{c}{Common Sense} & \multicolumn{2}{c}{Symbolic Reasoning} & \multicolumn{2}{c}{Other  Tasks}  \\ 
\cmidrule(r){2-3}%
\cmidrule(r){4-5}%
\cmidrule(r){6-7}%

% \cline{2-7}
                        & CSQA       & Strategy            & Letter & Coin                          & Date     & Object                          \\ 
\midrule
Few-Shot-CoT               & 76.3       & 67.4                & 74.2   & 99.6                          & 78.9     & 56.7                            \\ 
\midrule
Zero-Shot               & 74.5       & 66.0                & 23.8   & 55.2                          & 67.8     & 38.7                            \\ 
Zero-Shot-CoT               & 68.8       & 65.8                & 53.2   & \textbf{98.8}                          & 65.9     & \textbf{73.5}                            \\
Role-Play Prompting     & \textbf{77.2}       & \textbf{67.0}                & \textbf{84.2}   & 89.4                          & \textbf{69.9}     & 67.7                            \\
\midrule
CoT in Zero-Shot &\checkmark  &\checkmark &\ding{55} &\ding{55} &\checkmark &\ding{55} \\
\bottomrule
\end{tabular}}
\caption{Accuracy comparison of Role-Play Prompting with Few-Shot-CoT,  Zero-Shot, Zero-Shot-CoT on each dataset. In the rows ``CoT in Zero-Shot", the check mark denotes that ChatGPT can spontaneously generate CoT on the corresponding dataset under the zero-shot setting, while the cross (wrong symbol) denotes otherwise.}
\label{tb: results}
\end{table*}

\begin{table*}[!h]
\centering
\scalebox{0.95}{
\begin{tabular}{p{15.5cm}}
\toprule
\textbf{Question}: Take the last letters of each words in ``Williams Reza Ashton Lillian" and concatenate them. \\
\midrule
\textbf{Zero-Shot}: saznln \quad\textcolor{red}{\ding{55}} \\
\midrule
\textbf{Zero-Shot-CoT}: \\Step 1: Write down the words - ``Williams Reza Ashton Lillian" \\Step 2: Take the last letter of each word - ``s a n n n" \\Step 3: Concatenate the last letters - ``sannn" \\So, the final result is ``sannn". \quad\textcolor{red}{\ding{55}}                                             \\ 
\midrule
\textbf{Role-Play Prompting}:\\ Sure! To create a new word by concatenating the last letters of each word in ``Williams Reza Ashton Lillian," we'll take the last letter of each word and combine them. Let's do it step by step: \\1. Williams: The last letter is ``s." \\2. Reza: The last letter is ``a." \\3. Ashton: The last letter is ``n." \\4. Lillian: The last letter is ``n." \\Now, let's concatenate these last letters: ``s" + ``a" + ``n" + ``n" = ``sann." \\Therefore, by concatenating the last letters of each word, we get the new word ``sann."\quad\textcolor{green!80!black}{\checkmark}\\
\bottomrule
\end{tabular}}
\caption{An example of Zero-Shot, Zero-Shot-CoT, and Role-Play Prompting on Last Letter Concatenation.}
\label{tb: letter example}
\end{table*}

\noindent{\bf Model}\quad
We use ChatGPT (gpt-3.5-turbo-0613), the current strongest conversational model in addition to GPT-4, to conduct experiments. % \cite{gpt4}

\noindent{\bf Prompt}\quad
Our approach involves the design of a role-setting prompt and a role-feedback prompt for a given task. The arithmetic task consists of six datasets, all utilizing the same prompts, as depicted in Figure \ref{fg: example}. Similarly, the common sense reasoning task comprises two datasets, also employing the same prompts as shown in Figure \ref{fg: step}. For other tasks, the prompts used are detailed in Table \ref{tb: propmts}.

\noindent{\bf Baselines}\quad
We choose the standard zero-shot prompting, Zero-Shot-CoT \cite{zero_shot_cot}, and Few-Shot-CoT \cite{chain} as baselines. 
Following previous work \cite{zero_shot_cot,auto_cot}, we use greedy decoding for all the experiments by setting the temperature to $0$, making the results deterministic. See more details in Appendix \ref{app: baselines}.

\subsection{Results and Analysis}

\begin{table*}
\centering
\scalebox{0.82}{
\begin{tabular}{c|p{16.5cm}|c} 
\toprule
No.             &  \multicolumn{1}{c}{Prompt}                                                                                                                                                                                                                                                                                                                      & AQuA               \\ 
\midrule
1                  & From now on, you are a math teacher. Please answer
  the following question.                                                                                                                                                                                                                           & 57.1                   \\  
\midrule
\multirow{1}{*}[-0.7em]{2}                  & From now on, you are an \colorbox{gray!17}{excellent} math teacher and
  always teach your students math problems \colorbox{gray!17}{correctly}. And \colorbox{gray!17}{I am one of your
  students} and ask you the following question.                                                                                                                                                & \multirow{1}{*}[-0.7em]{60.2}                   \\ 
\midrule
\midrule
\multirow{2}{*}[-1.4em]{3} & From now on, you are an excellent math teacher and
  always teach your students math problems correctly. And I am one of your
  students.                                                                                                                                                                                   & \multirow{2}{*}[-1.4em]{61.4}  \\ 
\cmidrule{2-2}
                   & \colorbox{gray!17}{That's great to hear! As your math teacher, I'll do
  my best to explain mathematical concepts correctly so} \colorbox{gray!17}{that you can understand
  them easily. Feel free to ask any math problems or questions you have, and I'll be} \colorbox{gray!17}{glad to assist you.}                                                                              &                        \\ 
\midrule
\multirow{2}{*}[-1.4em]{4} & From now on, you are an excellent math teacher and
  always teach your students math problems correctly. And I am one of your
  students.                                                                                                                                                                                   & \multirow{2}{*}[-1.4em]{63.8}  \\ 
\cmidrule{2-2}
                   & That's great to hear! As your math teacher, I'll do
  my best to explain mathematical concepts correctly so that you can understand
  them easily. Feel free to ask any math problems or questions you have, and
  I'll be glad to assist you. \colorbox{gray!17}{Let's dive into the world of mathematics and
  explore its wonders together!} &                        \\
\bottomrule
\end{tabular}}
\caption{Accuracy comparison of different prompt designs with a fixed role of the math teacher on AQuA. We utilize gray shading to indicate the additional content in comparison to the previous prompt. Supplementary experiments in Appendix \ref{app: length effect} eliminate the possibility of performance increase caused by the increase in prompt length.}
\label{tb: internal design}
\end{table*}

\begin{table}
\centering
\scalebox{0.79}{
\begin{tabular}{ccccc} 
\toprule
No. & Category                       & Role             & AQuA & SVAMP  \\ 
\midrule
1   & \multirow{2}{*}{advantaged}    & math teacher     & 63.8 & 83.8   \\
2   &                                & mathematician    & 60.2 & 82.3   \\ 
\midrule
3   & \multirow{4}{*}{irrelevant}    & police           & 59.8 & 82.3   \\
4   &                                & farmer           & 59.8 & 82.2   \\
5   &                                & doctor           & 56.3 & 74.4   \\
6   &                                & writer           & 55.5 & 82.0   \\ 
\midrule
7   & \multirow{2}{*}{disadvantaged} & careless student & 51.6 & 68.7   \\
8   &                                & math rookie       & 45.3 & 72.9      \\
\bottomrule
\end{tabular}}
\caption{Accuracy comparison of different roles for role-play prompting on AQuA and SVAMP.}
\label{tb: role selection}
\end{table}

Comprehensive evaluation results are presented in Table \ref{tb: results}. The evaluation metric is accuracy. 

\noindent{\bf Comparison with Standard Zero-Shot}\quad
As shown in Table \ref{tb: results}, our role-play prompting approach demonstrates superior performance, outperforming the zero-shot baseline in \textbf{10 out of 12} datasets, and achieving on par performance in the remaining 2 datasets (SingleEq and MultiArith).
% Notably, it excels in $4$ out of $6$ arithmetic reasoning datasets, and all datasets of commonsense reasoning, symbolic reasoning, and other reasoning tasks from Big-bench. These substantial improvements strongly demonstrate the effectiveness of role-play prompting.
%We observe a slight underperformance of role-play prompting on SingleEq and MultiArith. We attribute this mainly to the high accuracy already achieved by these two datasets under the zero-shot setting, both surpassing 97\%. Given the relative simplicity of these datasets, the model's performance may have reached saturation, making it challenging for our method to further enhance accuracy at this high level. Despite a marginal shortfall in these two specific datasets, we emphasize the competitiveness of role-play prompting across a variety of more complex datasets. This strongly illustrates the effectiveness of role-play prompting in a wider range of application scenarios.
%A slight underperformance of role-play prompting is observed in the SingleEq and MultiArith datasets. This can primarily be attributed to the exceptionally high accuracy already attained in these two datasets under the zero-shot setting, where both exceed 97\%. 
Considering the relative simplicity of the SingleEq and MultiArith datasets, it is plausible that the model's performance has approached a saturation point (exceed 97\%), thereby presenting a significant challenge for our method to further enhance accuracy at such an elevated level. While achieving on par performance in these specific datasets, it is crucial to highlight the competitive nature of role-play prompting across a diverse array of more complex datasets. This strongly demonstrates the effectiveness of role-play prompting in an extensive range of application scenarios.

\noindent{\bf Comparison with Zero-Shot-CoT}\quad
Zero-Shot-CoT appends \emph{``Let's think step by step''} to the question to stimulate the chain of thought (CoT) in LLMs, making it a simple yet effective method to enhance the reasoning ability of LLMs. However, different from the earlier instructed LLMs \cite{hlrf}, the current conversational LLMs have undergone extensive supervised fine-tuning, which enables them to spontaneously generate CoT in some fields under the zero-shot setting. In this context, we conduct a comparative analysis of our role-play prompting approach with Zero-Shot-CoT. The experimental results, along with the model's ability to spontaneously generate CoT are presented in Table \ref{tb: results}. Note that the direct output of answers or a slight reasoning process is not considered CoT. Overall, our approach outperforms Zero-Shot-CoT on \textbf{9 out of 12} datasets. In tasks (Letter, Coin, Object) where ChatGPT struggles to generate CoT spontaneously, both of them gain huge improvements. Through the case study, we find that role-play prompting also stimulates CoT in the model just like Zero-Shot-CoT. An example is provided in Table \ref{tb: letter example}. In more tasks where CoT already occurs, both our approach and Zero-Shot-CoT reinforce the step-by-step reasoning process (examples are provided in Appendix \ref{app: svamp}). However, Zero-Shot-CoT demonstrates no significant effect while role-play prompting leads to better results. Therefore, we posit that role-play prompting serves as an implicit CoT trigger and can generate a more effective CoT.

\noindent{\bf Comparison with Few-Shot-CoT}\quad
Though our role-play prompting approach is completely zero-shot, the improvement it brings is nearly on par with Few-Shot-CoT, even surpassing Few-Shot-CoT on \textbf{6 out of 12} datasets.

Following previous work \cite{zero_shot_cot, plan-and-solve}, we combine our approach and baselines with Self-Consistency to further prove the efficacy of role-play prompting. Related results and discussions are provided in Appendix \ref{app: sc}. 

\subsection{Impact of Prompt Design}
\label{sec: prompt_design}

\begin{table*}
\centering
\scalebox{0.9}{
\begin{tabular}{llccccc} 
\toprule
Model                        & Method                & GSM8K & MultiArith & SVAMP & CSQA & Letter  \\ 
\midrule
\multirow{2}{*}{Llama-2-70B-Chat} & Zero-Shot             & 53.9  & 86.0       & 78.9   & - & 18.8    \\
                             & Role-Play
  Prompting & \textbf{58.9}  & \textbf{90.2}    & \textbf{79.0}   & -  & \textbf{25.8}    \\ 
\midrule
\multirow{2}{*}{Vicuna-33B}      & Zero-Shot             & 42.9  & 70.7       & \textbf{59.1}    &65.5  & 2.2     \\
                             & Role-Play
  Prompting & \textbf{44.9}  & \textbf{71.5}          & 58.5  & \textbf{67.2}     & \textbf{5.2}     \\
\bottomrule
\end{tabular}}
\caption{Accuracy comparison of Role-Play Prompting with Zero-Shot on open-source conversational LLMs. Due to safety concerns, Llama 2-Chat refuses to answer on CSQA, so the relevant results are not shown. See more details in Appendix \ref{app: more_llm}.}
\label{tb: other models}
\end{table*}

\begin{figure*}
\centering
\vspace{-0.2cm}
\subfigure[GSM8K]{\includegraphics[width=0.3\textwidth,trim=4cm 2cm 4cm 2cm, clip]{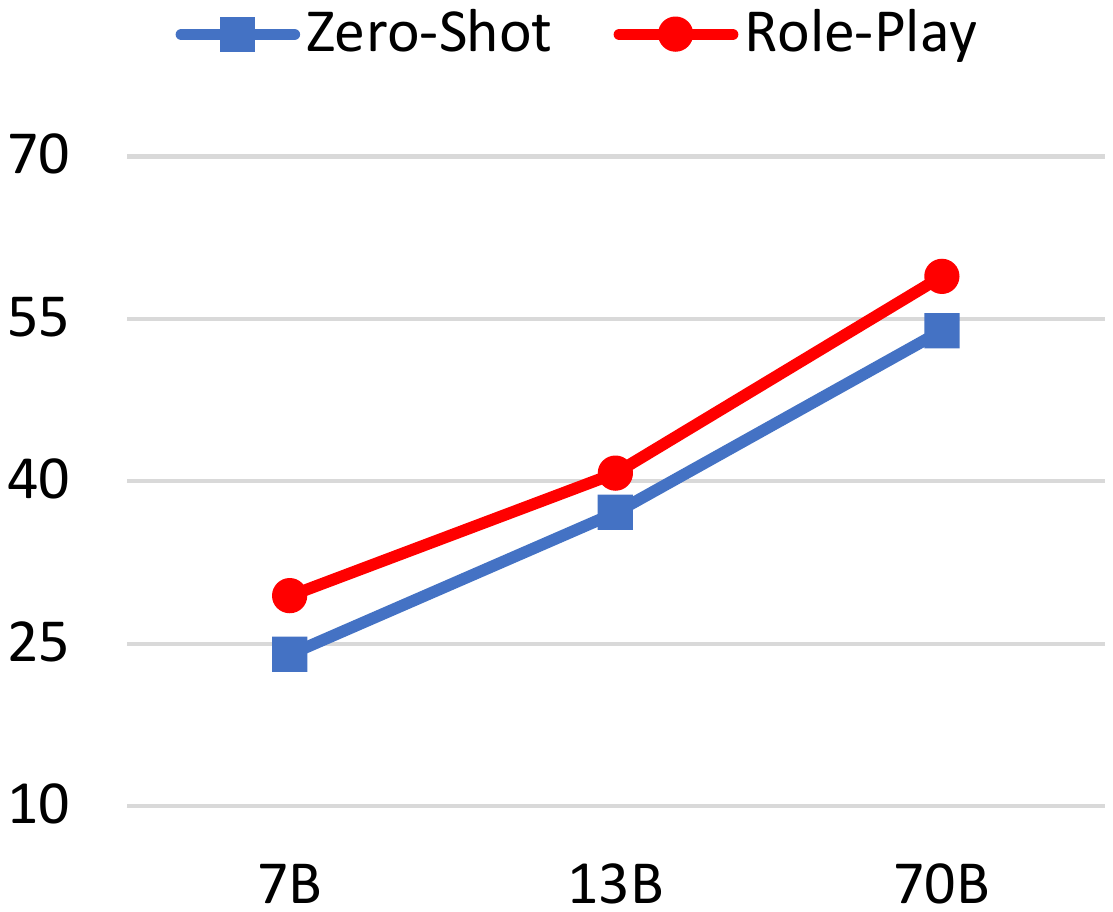}}
\subfigure[MultiArith]{\includegraphics[width=0.3\textwidth,trim=4cm 2cm 4cm 2cm, clip]{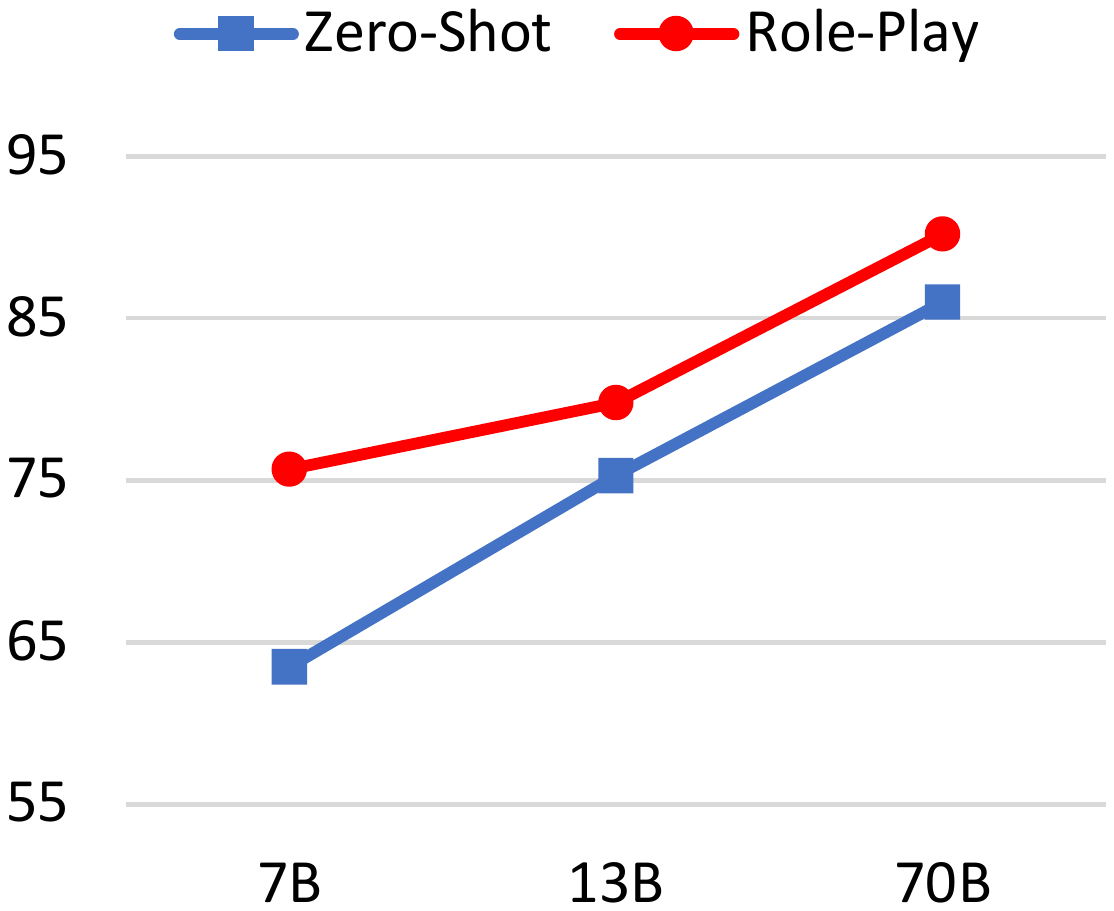}}
\subfigure[Letter]{\includegraphics[width=0.3\textwidth,trim=4cm 2cm 4cm 2cm, clip]{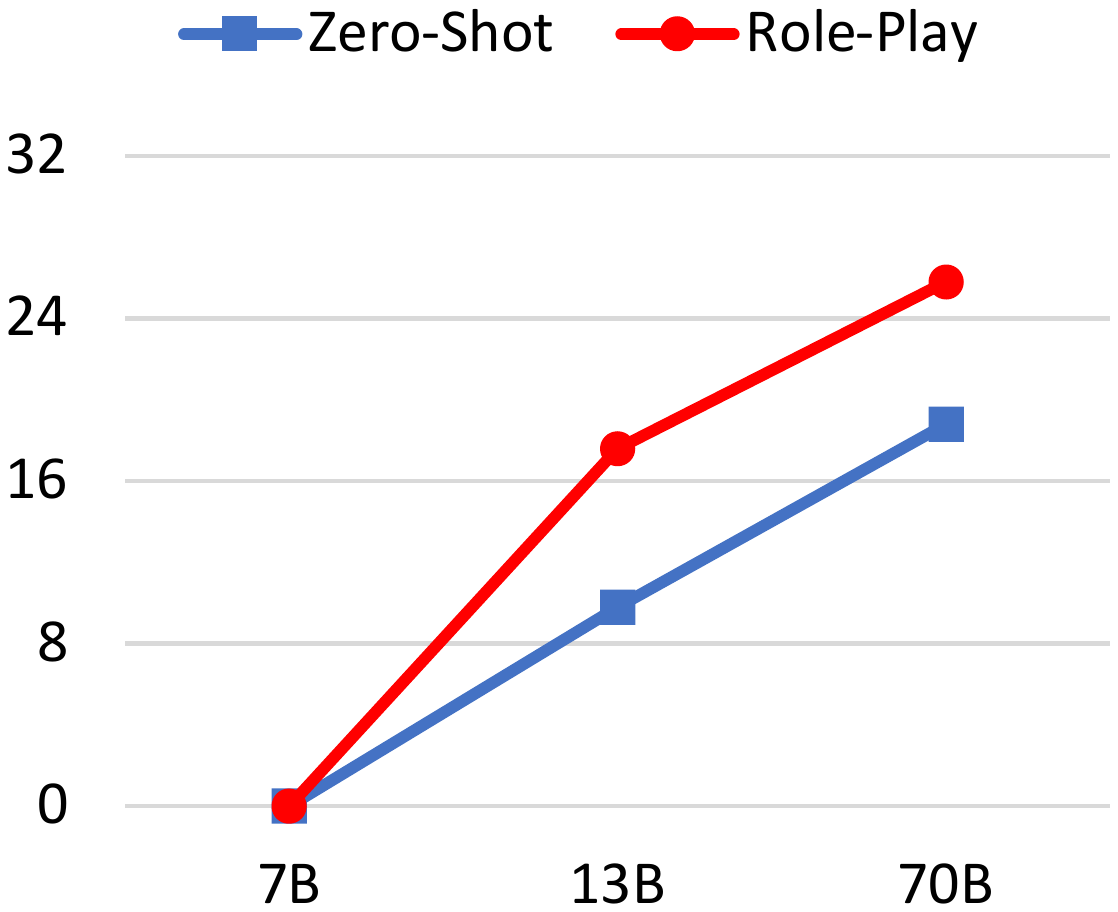}}
\vspace{-0.1cm}
\caption{Accuracy comparison of Role-Play Prompting across different sizes of Llama 2-Chat models. See more details in Appendix \ref{app: model_scale}.} %图片标题
\label{fg: other sizes}  %图片交叉引用时的标签
\end{figure*}

\noindent{\bf Prompt Structure}\quad
To determine the optimal prompt structure, 
%for maximizing role immersion,
we select AQuA dataset and assign the model the role of a math teacher. We then conduct ablation studies on this setup to systematically assess the impact of different design choices. We hypothesize that prompts which immerse the model deeper in its role will improve performance. Consequently, we design four groups of prompts with progressively increasing levels of immersion, as shown in Table \ref{tb: internal design}. Prompt 1 and 2 are designed as single-round dialogues, where we directly attach the question to the prompt and input it into the model to obtain the answer. Prompt 1 solely contains the role to be played, and it already achieves the result surpassing the zero-shot baseline. For Prompt 2, we further enhance immersion by adding complementary descriptions of the role and specifying relevant roles for the user. This enhancement further improve the performance. Prompt 3 and 4 are both designed as two-round dialogues, as described in the previous section. By allowing the model to respond to the given role setting, the immersion is further enhanced, leading to the best performance. We conduct the same experiments on Letter and Coin datasets, yielding consistent findings (see more details in Appendix \ref{app: letter}). Therefore, we recommend using the two-round prompt structure with complementary descriptions to maximize the model's immersion, thereby unlocking the full reasoning potential of role-play prompting.

\noindent{\bf Role Selection}\quad
To assess the impact of role selection, we test on the AQuA and SVAMP arithmetic datasets using two-round dialogue prompts. We design 8 varied roles, categorized as advantaged, irrelevant, or disadvantaged based on whether each role holds an advantage in the given task. The performance of these roles is detailed in Table \ref{tb: role selection}, while the specific prompt designs can be found in Appendix \ref{app: prompt}. Consistent with intuition, advantaged roles (1,2) undoubtedly achieve the best results, followed by irrelevant roles (3-6) (surprisingly, most of them outperform the zero-shot baseline even though they have no advantage on arithmetic tasks), and disadvantaged roles (7,8) achieve the worst results, underperforming the zero-shot baseline. Therefore, we recommend choosing a role that holds an advantage in the given task for role-play prompting.

\subsection{Experiments on More LLMs}

To assess the generalization of our role-play prompting approach, we conduct additional experiments using several open-source conversational LLMs, including Llama 2-Chat \cite{llama-2} and Vicuna \cite{vicuna2023}, on various datasets such as GSM8K, MultiArith, SVAMP, CSQA, and Letter. The prompts and the decoding strategy used are consistent with the previous ChatGPT experiments. The results are shown in Table \ref{tb: other models}, which indicate that role-play prompting also exceeds the zero-shot baseline in open-source conversational LLMs, demonstrating the good generalization ability of role-play prompting.

Furthermore, we examine the impact of model scale by testing the Llama 2-Chat series (7B, 13B, 70B) on GSM8K, MultiArith, and Letter datasets. As Figure \ref{fg: other sizes} illustrates, all three model sizes achieve improved performance from role-play prompting. The consistent benefits across 7B to 70B parameters indicate efficacy independent of scale, within this range.

\section{Conclusion}

In this paper, we have proposed a novel zero-shot role-play prompting methodology consisting of a two-stage framework, aimed at enhancing the reasoning capabilities of LLMs. Extensive evaluations across twelve widely-used benchmarks reveal that our approach outperforms both the standard zero-shot baseline and Zero-Shot-CoT on most of the datasets. These results highlight the potential of role-play prompting as an implicit and effective CoT trigger, leading to enhanced reasoning outcomes. Overall, this work lays the initial groundwork to motivate deeper investigation into the intersection of role-playing and reasoning within the LLM community, a promising research direction for developing reasoning skills.

\section*{Limitations}

The core of our role-play prompting approach lies in the design of the role-setting and role-feedback prompts. While we have manually designed and sampled some prompts, yielding superior results compared to the zero-shot baseline, this process is time-consuming and may not always guarantee optimal results. To address this limitation, future research could focus on enabling LLMs to autonomously choose appropriate roles and design prompts based on the given question. This approach could further extend the application of role-play prompting to a broader range of domains beyond reasoning.

\section*{Acknowledgements}

The work was supported by National Key R\&D Program of China (No.2022ZD0116307), National Natural Science Foundation of China (No. 62271270) and Sponsored by CCF-Lenovo Bule Ocean Research Fund.

\bibliography{anthology,custom}

\begin{thebibliography}{30}
\expandafter\ifx\csname natexlab\endcsname\relax\def\natexlab#1{#1}\fi

\bibitem[{Brown et~al.(2020)Brown, Mann, Ryder, Subbiah, Kaplan, Dhariwal, Neelakantan, Shyam, Sastry, Askell, Agarwal, Herbert-Voss, Krueger, Henighan, Child, Ramesh, Ziegler, Wu, Winter, Hesse, Chen, Sigler, Litwin, Gray, Chess, Clark, Berner, McCandlish, Radford, Sutskever, and Amodei}]{NEURIPS2020_1457c0d6}
Tom Brown, Benjamin Mann, Nick Ryder, Melanie Subbiah, Jared~D Kaplan, Prafulla Dhariwal, Arvind Neelakantan, Pranav Shyam, Girish Sastry, Amanda Askell, Sandhini Agarwal, Ariel Herbert-Voss, Gretchen Krueger, Tom Henighan, Rewon Child, Aditya Ramesh, Daniel Ziegler, Jeffrey Wu, Clemens Winter, Chris Hesse, Mark Chen, Eric Sigler, Mateusz Litwin, Scott Gray, Benjamin Chess, Jack Clark, Christopher Berner, Sam McCandlish, Alec Radford, Ilya Sutskever, and Dario Amodei. 2020.
\newblock \href {https://proceedings.neurips.cc/paper_files/paper/2020/file/1457c0d6bfcb4967418bfb8ac142f64a-Paper.pdf} {Language models are few-shot learners}.
\newblock In \emph{Advances in Neural Information Processing Systems}, volume~33, pages 1877--1901. Curran Associates, Inc.

\bibitem[{Chiang et~al.(2023)Chiang, Li, Lin, Sheng, Wu, Zhang, Zheng, Zhuang, Zhuang, Gonzalez, Stoica, and Xing}]{vicuna2023}
Wei-Lin Chiang, Zhuohan Li, Zi~Lin, Ying Sheng, Zhanghao Wu, Hao Zhang, Lianmin Zheng, Siyuan Zhuang, Yonghao Zhuang, Joseph~E. Gonzalez, Ion Stoica, and Eric~P. Xing. 2023.
\newblock \href {https://lmsys.org/blog/2023-03-30-vicuna/} {Vicuna: An open-source chatbot impressing gpt-4 with 90\%* chatgpt quality}.

\bibitem[{Chowdhery et~al.(2022)Chowdhery, Narang, Devlin, Bosma, Mishra, Roberts, Barham, Chung, Sutton, Gehrmann et~al.}]{chowdhery2022palm}
Aakanksha Chowdhery, Sharan Narang, Jacob Devlin, Maarten Bosma, Gaurav Mishra, Adam Roberts, Paul Barham, Hyung~Won Chung, Charles Sutton, Sebastian Gehrmann, et~al. 2022.
\newblock Palm: Scaling language modeling with pathways.
\newblock \emph{arXiv preprint arXiv:2204.02311}.

\bibitem[{Cobbe et~al.(2021)Cobbe, Kosaraju, Bavarian, Chen, Jun, Kaiser, Plappert, Tworek, Hilton, Nakano et~al.}]{gsm8k}
Karl Cobbe, Vineet Kosaraju, Mohammad Bavarian, Mark Chen, Heewoo Jun, Lukasz Kaiser, Matthias Plappert, Jerry Tworek, Jacob Hilton, Reiichiro Nakano, et~al. 2021.
\newblock Training verifiers to solve math word problems.
\newblock \emph{arXiv preprint arXiv:2110.14168}.

\bibitem[{Diao et~al.(2023)Diao, Wang, Lin, and Zhang}]{activeprompt}
Shizhe Diao, Pengcheng Wang, Yong Lin, and Tong Zhang. 2023.
\newblock Active prompting with chain-of-thought for large language models.
\newblock \emph{arXiv preprint arXiv:2302.12246}.

\bibitem[{Geva et~al.(2021)Geva, Khashabi, Segal, Khot, Roth, and Berant}]{strategy}
Mor Geva, Daniel Khashabi, Elad Segal, Tushar Khot, Dan Roth, and Jonathan Berant. 2021.
\newblock \href {https://doi.org/10.1162/tacl_a_00370} {Did aristotle use a laptop? a question answering benchmark with implicit reasoning strategies}.
\newblock \emph{Transactions of the Association for Computational Linguistics}, 9:346--361.

\bibitem[{Han et~al.(2022)Han, Kim, Yoo, Seo, Kim, Erdenee, and Chang}]{han-etal-2022-meet}
Seungju Han, Beomsu Kim, Jin~Yong Yoo, Seokjun Seo, Sangbum Kim, Enkhbayar Erdenee, and Buru Chang. 2022.
\newblock \href {https://doi.org/10.18653/v1/2022.naacl-main.377} {Meet your favorite character: Open-domain chatbot mimicking fictional characters with only a few utterances}.
\newblock In \emph{Proceedings of the 2022 Conference of the North American Chapter of the Association for Computational Linguistics: Human Language Technologies}, pages 5114--5132, Seattle, United States. Association for Computational Linguistics.

\bibitem[{Hosseini et~al.(2014)Hosseini, Hajishirzi, Etzioni, and Kushman}]{addsub}
Mohammad~Javad Hosseini, Hannaneh Hajishirzi, Oren Etzioni, and Nate Kushman. 2014.
\newblock \href {https://doi.org/10.3115/v1/D14-1058} {Learning to solve arithmetic word problems with verb categorization}.
\newblock In \emph{Proceedings of the 2014 Conference on Empirical Methods in Natural Language Processing ({EMNLP})}, pages 523--533, Doha, Qatar. Association for Computational Linguistics.

\bibitem[{Kojima et~al.(2022)Kojima, Gu, Reid, Matsuo, and Iwasawa}]{zero_shot_cot}
Takeshi Kojima, Shixiang~(Shane) Gu, Machel Reid, Yutaka Matsuo, and Yusuke Iwasawa. 2022.
\newblock \href {https://proceedings.neurips.cc/paper_files/paper/2022/file/8bb0d291acd4acf06ef112099c16f326-Paper-Conference.pdf} {Large language models are zero-shot reasoners}.
\newblock In \emph{Advances in Neural Information Processing Systems}, volume~35, pages 22199--22213. Curran Associates, Inc.

\bibitem[{Koncel-Kedziorski et~al.(2015)Koncel-Kedziorski, Hajishirzi, Sabharwal, Etzioni, and Ang}]{singleeq}
Rik Koncel-Kedziorski, Hannaneh Hajishirzi, Ashish Sabharwal, Oren Etzioni, and Siena~Dumas Ang. 2015.
\newblock \href {https://doi.org/10.1162/tacl_a_00160} {Parsing algebraic word problems into equations}.
\newblock \emph{Transactions of the Association for Computational Linguistics}, 3:585--597.

\bibitem[{Li et~al.(2023)Li, Lin, Zhang, Fu, Chen, Lou, and Chen}]{diverse}
Yifei Li, Zeqi Lin, Shizhuo Zhang, Qiang Fu, Bei Chen, Jian-Guang Lou, and Weizhu Chen. 2023.
\newblock \href {https://aclanthology.org/2023.acl-long.291} {Making language models better reasoners with step-aware verifier}.
\newblock In \emph{Proceedings of the 61st Annual Meeting of the Association for Computational Linguistics (Volume 1: Long Papers)}, pages 5315--5333, Toronto, Canada. Association for Computational Linguistics.

\bibitem[{Ling et~al.(2017)Ling, Yogatama, Dyer, and Blunsom}]{aqua}
Wang Ling, Dani Yogatama, Chris Dyer, and Phil Blunsom. 2017.
\newblock \href {https://doi.org/10.18653/v1/P17-1015} {Program induction by rationale generation: Learning to solve and explain algebraic word problems}.
\newblock In \emph{Proceedings of the 55th Annual Meeting of the Association for Computational Linguistics (Volume 1: Long Papers)}, pages 158--167, Vancouver, Canada. Association for Computational Linguistics.

\bibitem[{Madaan et~al.(2023)Madaan, Tandon, Gupta, Hallinan, Gao, Wiegreffe, Alon, Dziri, Prabhumoye, Yang et~al.}]{selfrefine}
Aman Madaan, Niket Tandon, Prakhar Gupta, Skyler Hallinan, Luyu Gao, Sarah Wiegreffe, Uri Alon, Nouha Dziri, Shrimai Prabhumoye, Yiming Yang, et~al. 2023.
\newblock Self-refine: Iterative refinement with self-feedback.
\newblock \emph{arXiv preprint arXiv:2303.17651}.

\bibitem[{Ouyang et~al.(2022)Ouyang, Wu, Jiang, Almeida, Wainwright, Mishkin, Zhang, Agarwal, Slama, Ray, Schulman, Hilton, Kelton, Miller, Simens, Askell, Welinder, Christiano, Leike, and Lowe}]{hlrf}
Long Ouyang, Jeffrey Wu, Xu~Jiang, Diogo Almeida, Carroll Wainwright, Pamela Mishkin, Chong Zhang, Sandhini Agarwal, Katarina Slama, Alex Ray, John Schulman, Jacob Hilton, Fraser Kelton, Luke Miller, Maddie Simens, Amanda Askell, Peter Welinder, Paul~F Christiano, Jan Leike, and Ryan Lowe. 2022.
\newblock \href {https://proceedings.neurips.cc/paper_files/paper/2022/file/b1efde53be364a73914f58805a001731-Paper-Conference.pdf} {Training language models to follow instructions with human feedback}.
\newblock In \emph{Advances in Neural Information Processing Systems}, volume~35, pages 27730--27744. Curran Associates, Inc.

\bibitem[{Patel et~al.(2021)Patel, Bhattamishra, and Goyal}]{svamp}
Arkil Patel, Satwik Bhattamishra, and Navin Goyal. 2021.
\newblock \href {https://doi.org/10.18653/v1/2021.naacl-main.168} {Are {NLP} models really able to solve simple math word problems?}
\newblock In \emph{Proceedings of the 2021 Conference of the North American Chapter of the Association for Computational Linguistics: Human Language Technologies}, pages 2080--2094, Online. Association for Computational Linguistics.

\bibitem[{Rae et~al.(2021)Rae, Borgeaud, Cai, Millican, Hoffmann, Song, Aslanides, Henderson, Ring, Young et~al.}]{rae2021scaling}
Jack~W Rae, Sebastian Borgeaud, Trevor Cai, Katie Millican, Jordan Hoffmann, Francis Song, John Aslanides, Sarah Henderson, Roman Ring, Susannah Young, et~al. 2021.
\newblock Scaling language models: Methods, analysis \& insights from training gopher.
\newblock \emph{arXiv preprint arXiv:2112.11446}.

\bibitem[{Roy and Roth(2015)}]{multiarith}
Subhro Roy and Dan Roth. 2015.
\newblock \href {https://doi.org/10.18653/v1/D15-1202} {Solving general arithmetic word problems}.
\newblock In \emph{Proceedings of the 2015 Conference on Empirical Methods in Natural Language Processing}, pages 1743--1752, Lisbon, Portugal. Association for Computational Linguistics.

\bibitem[{Shanahan et~al.(2023)Shanahan, McDonell, and Reynolds}]{shanahan2023role}
Murray Shanahan, Kyle McDonell, and Laria Reynolds. 2023.
\newblock Role-play with large language models.
\newblock \emph{arXiv preprint arXiv:2305.16367}.

\bibitem[{Srivastava et~al.(2022)Srivastava, Rastogi, Rao, Shoeb, Abid, Fisch, Brown, Santoro, Gupta, Garriga-Alonso et~al.}]{srivastava2022beyond}
Aarohi Srivastava, Abhinav Rastogi, Abhishek Rao, Abu Awal~Md Shoeb, Abubakar Abid, Adam Fisch, Adam~R Brown, Adam Santoro, Aditya Gupta, Adri{\`a} Garriga-Alonso, et~al. 2022.
\newblock Beyond the imitation game: Quantifying and extrapolating the capabilities of language models.
\newblock \emph{arXiv preprint arXiv:2206.04615}.

\bibitem[{Talmor et~al.(2019)Talmor, Herzig, Lourie, and Berant}]{csqa}
Alon Talmor, Jonathan Herzig, Nicholas Lourie, and Jonathan Berant. 2019.
\newblock \href {https://doi.org/10.18653/v1/N19-1421} {{C}ommonsense{QA}: A question answering challenge targeting commonsense knowledge}.
\newblock In \emph{Proceedings of the 2019 Conference of the North {A}merican Chapter of the Association for Computational Linguistics: Human Language Technologies, Volume 1 (Long and Short Papers)}, pages 4149--4158, Minneapolis, Minnesota. Association for Computational Linguistics.

\bibitem[{Touvron et~al.(2023{\natexlab{a}})Touvron, Lavril, Izacard, Martinet, Lachaux, Lacroix, Rozi{\`e}re, Goyal, Hambro, Azhar et~al.}]{touvron2023llama}
Hugo Touvron, Thibaut Lavril, Gautier Izacard, Xavier Martinet, Marie-Anne Lachaux, Timoth{\'e}e Lacroix, Baptiste Rozi{\`e}re, Naman Goyal, Eric Hambro, Faisal Azhar, et~al. 2023{\natexlab{a}}.
\newblock Llama: Open and efficient foundation language models.
\newblock \emph{arXiv preprint arXiv:2302.13971}.

\bibitem[{Touvron et~al.(2023{\natexlab{b}})Touvron, Martin, Stone, Albert, Almahairi, Babaei, Bashlykov, Batra, Bhargava, Bhosale et~al.}]{llama-2}
Hugo Touvron, Louis Martin, Kevin Stone, Peter Albert, Amjad Almahairi, Yasmine Babaei, Nikolay Bashlykov, Soumya Batra, Prajjwal Bhargava, Shruti Bhosale, et~al. 2023{\natexlab{b}}.
\newblock Llama 2: Open foundation and fine-tuned chat models.
\newblock \emph{arXiv preprint arXiv:2307.09288}.

\bibitem[{Wan et~al.(2023)Wan, Sun, Dai, Arik, and Pfister}]{COSP}
Xingchen Wan, Ruoxi Sun, Hanjun Dai, Sercan Arik, and Tomas Pfister. 2023.
\newblock \href {https://aclanthology.org/2023.findings-acl.216} {Better zero-shot reasoning with self-adaptive prompting}.
\newblock In \emph{Findings of the Association for Computational Linguistics: ACL 2023}, pages 3493--3514, Toronto, Canada. Association for Computational Linguistics.

\bibitem[{Wang et~al.(2023)Wang, Xu, Lan, Hu, Lan, Lee, and Lim}]{plan-and-solve}
Lei Wang, Wanyu Xu, Yihuai Lan, Zhiqiang Hu, Yunshi Lan, Roy Ka-Wei Lee, and Ee-Peng Lim. 2023.
\newblock \href {https://doi.org/10.18653/v1/2023.acl-long.147} {Plan-and-solve prompting: Improving zero-shot chain-of-thought reasoning by large language models}.
\newblock In \emph{Proceedings of the 61st Annual Meeting of the Association for Computational Linguistics (Volume 1: Long Papers)}, pages 2609--2634, Toronto, Canada. Association for Computational Linguistics.

\bibitem[{Wang et~al.(2022)Wang, Wei, Schuurmans, Le, Chi, Narang, Chowdhery, and Zhou}]{selfcon}
Xuezhi Wang, Jason Wei, Dale Schuurmans, Quoc~V Le, Ed~H Chi, Sharan Narang, Aakanksha Chowdhery, and Denny Zhou. 2022.
\newblock Self-consistency improves chain of thought reasoning in language models.
\newblock In \emph{The Eleventh International Conference on Learning Representations}.

\bibitem[{Wei et~al.(2022)Wei, Wang, Schuurmans, Bosma, ichter, Xia, Chi, Le, and Zhou}]{chain}
Jason Wei, Xuezhi Wang, Dale Schuurmans, Maarten Bosma, brian ichter, Fei Xia, Ed~Chi, Quoc~V Le, and Denny Zhou. 2022.
\newblock \href {https://proceedings.neurips.cc/paper_files/paper/2022/file/9d5609613524ecf4f15af0f7b31abca4-Paper-Conference.pdf} {Chain-of-thought prompting elicits reasoning in large language models}.
\newblock In \emph{Advances in Neural Information Processing Systems}, volume~35, pages 24824--24837. Curran Associates, Inc.

\bibitem[{Wu et~al.(2023)Wu, Gong, Shou, Liang, and Jiang}]{summarization}
Ning Wu, Ming Gong, Linjun Shou, Shining Liang, and Daxin Jiang. 2023.
\newblock Large language models are diverse role-players for summarization evaluation.
\newblock \emph{arXiv preprint arXiv:2303.15078}.

\bibitem[{Yao et~al.(2023)Yao, Yu, Zhao, Shafran, Griffiths, Cao, and Narasimhan}]{tot}
Shunyu Yao, Dian Yu, Jeffrey Zhao, Izhak Shafran, Thomas~L Griffiths, Yuan Cao, and Karthik Narasimhan. 2023.
\newblock Tree of thoughts: Deliberate problem solving with large language models.
\newblock \emph{arXiv preprint arXiv:2305.10601}.

\bibitem[{Zhang et~al.(2022)Zhang, Zhang, Li, and Smola}]{auto_cot}
Zhuosheng Zhang, Aston Zhang, Mu~Li, and Alex Smola. 2022.
\newblock Automatic chain of thought prompting in large language models.
\newblock In \emph{The Eleventh International Conference on Learning Representations}.

\bibitem[{Zhou et~al.(2022)Zhou, Sch{\"a}rli, Hou, Wei, Scales, Wang, Schuurmans, Cui, Bousquet, Le et~al.}]{leasttomost}
Denny Zhou, Nathanael Sch{\"a}rli, Le~Hou, Jason Wei, Nathan Scales, Xuezhi Wang, Dale Schuurmans, Claire Cui, Olivier Bousquet, Quoc~V Le, et~al. 2022.
\newblock Least-to-most prompting enables complex reasoning in large language models.
\newblock In \emph{The Eleventh International Conference on Learning Representations}.

\end{thebibliography}

\newpage
\appendix

\section{Implementation Details}

\subsection{Code for Calling ChatGPT’s API}
\label{app: call_api}

To help understand our approach of role-play prompting, we provide a code example of making an API call as follows. More details can be found in the API document \footnote{https://platform.openai.com/docs/api-reference/introduction} of OpenAI.

\lstset{    
     basicstyle = \scriptsize\ttfamily,             % 设置行距，字体
     numbers = none,                                      % 在左侧显示行号
     numberstyle = \footnotesize \color{darkgray},    
     xleftmargin = 0pt,
     commentstyle = \color[RGB]{0,96,96},                 % 设置代码注释的格式
     stringstyle = \color[RGB]{128,0,0},                  % 设置字符串格式
     frame =tb,                                      % 不显示背景边框
     backgroundcolor = \color{white},          % 设定背景颜色
     language=Python,                                           % 设置语言
     aboveskip = 10pt, % 调整上方间距
    belowskip = 10pt % 调整下方间距
}
\begin{lstlisting}
# A code example of making an API call   
prompt_1 = role_setting_prompt
prompt_2 = role_feedback_prompt
conversation = [
    {"role": "user", "content": prompt_1},
    {"role": "assistant", "content": prompt_2}, 
    {"role": "user", "content": question}
]
answer = openai.ChatCompletion.create(
    model="gpt-3.5-turbo-0613",
    messages=conversation,
    temperature=0,
    max_tokens=512
)
\end{lstlisting}

\subsection{Answer Extraction}

\begin{figure}[h]
\centering
\includegraphics[width=0.45\textwidth,trim=0cm 6cm 19cm 0cm, clip]{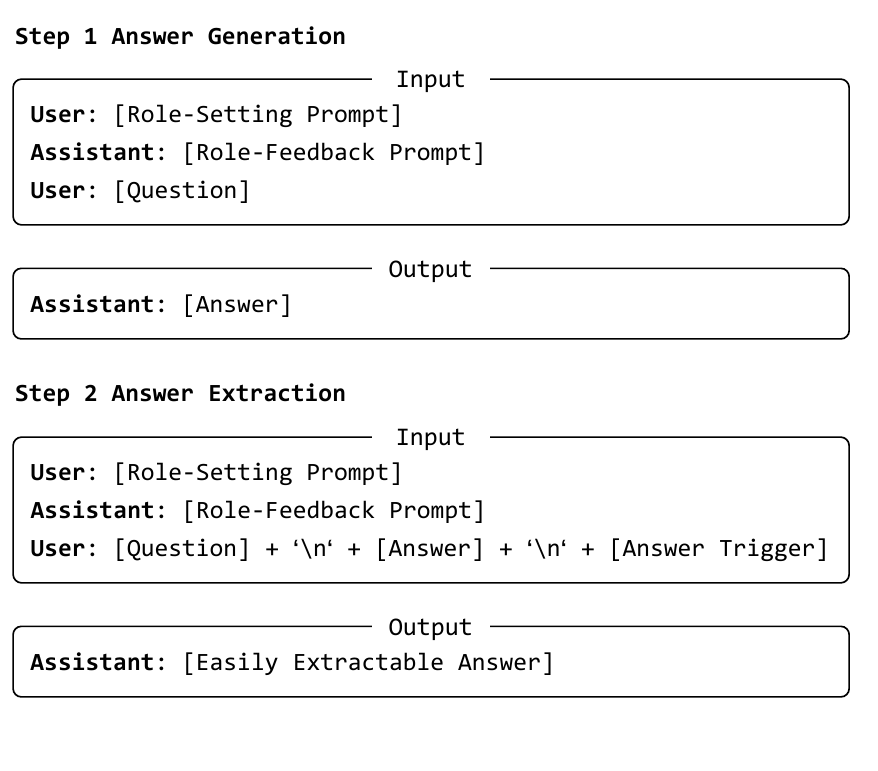} 
\vspace{-0.3cm}
\caption{A sketch map of answer extraction for role-play prompting.}
\label{fg: ans_extrac}
\end{figure}

Different from few-shot, the form of the answer given by LLMs under the zero-shot setting is not fixed. To simplify the extraction of answers, we follow the approach of Zero-Shot-CoT \cite{zero_shot_cot}. Specifically, for each question, after getting the answer generated by the LLM, we concatenate the question, answer, and answer trigger together and input them to the model. A sketch map of answer extraction for role-play prompting is shown in Figure \ref{fg: ans_extrac}. The answer trigger sentences for various answer formats are shown in Table \ref{tb: trigger}. More details can be found in the code.

\begin{table*}
\centering
% \scalebox{0.85}{
\begin{tabular}{ll} 
\toprule
Answer
  Format & Answer
  Trigger                               \\ 
\midrule
arabic number     & Therefore,
  the answer (arabic numerals) is   \\
option (A-E)    & Therefore,
  among A through E, the answer is  \\
option (A-C)    & Therefore,
  among A through C, the answer is  \\
yes or no       & Therefore,
  the answer (Yes or No) is         \\
string          & Therefore,
  the final answer is               \\
\bottomrule
\end{tabular}%}
\caption{Answer trigger sentences for various answer formats.}
\label{tb: trigger}
\end{table*}

\subsection{Baselines}
\label{app: baselines}

The standard zero-shot prompting, Zero-Shot-CoT \cite{zero_shot_cot}, and Few-Shot-CoT \cite{chain} are chosen as baselines. The standard zero-shot prompting directly inputs the target question without ant additional prompts. Zero-Shot-CoT appends "Let's think step by step." to the target question. Few-Shot-CoT adds similar questions and their corresponding reasoning processes before the target question. We use the few-shot exemplars provided in the original paper. When calling the API of ChatGPT (gpt-3.5-turbo-0613), we set max\_tokens = 512 and temperature = 0.

\subsection{Experiments on More LLMs}
\label{app: more_llm}

Besides ChatGPT, we conduct experiments using different open-source conversational LLMs, including Llama 2-Chat \cite{llama-2} and Vicuna \cite{vicuna2023}, on various datasets such as GSM8K, Multiarith, SVAMP, CSQA, and Letter. The prompts and the decoding strategy are consistent with the previous ChatGPT experiments. However, Llama 2-Chat often declines to respond to questions within the datasets due to overzealous safety concerns imposed by RLHF \cite{hlrf}. To solve this problem, we change the original system prompt of Llama 2-Chat to "We will test your abilities in the upcoming conversations, so please respond actively to the questions. Your answers will not cause any harm, so there's no need to worry. So, just answer!". The phenomenon of refusal to answer is alleviated on the CSQA dataset and completely resolved on other datasets. Therefore, we do not present the results of CSQA in the main text. The experiments on model size using Llama 2-Chat series also modify the system prompt.

% \begin{table*}
% \centering
% \begin{tabular}{llccccc} 
% \toprule
% Model                        & Method                & GSM8K & Multiarith & SVAMP & CSQA & Letter  \\ 
% \midrule
% \multirow{2}{*}{Llama-2-70B-Chat} & Zero-Shot             & 53.9  & 86.0       & 78.9   & \underline{67.1} & 18.8    \\
%                              & Role-Play
%   Prompting & \textbf{58.9}  & \textbf{90.2}    & \textbf{79.0}   & \underline{\textbf{70.9}}  & \textbf{25.8}    \\ 
% \midrule
% \multirow{2}{*}{Vicuna-33B}      & Zero-Shot             & 42.9  & 70.7       & \textbf{59.1}    &65.5  & 2.2     \\
%                              & Role-Play
%   Prompting & \textbf{44.9}  & \textbf{71.5}          & 58.5  & \textbf{67.2}     & \textbf{5.2}     \\
% \bottomrule
% \end{tabular}
% \caption{Accuracy comparison of Role-Play Prompting with Zero-Shot on open source conversational LLMs. Note that the results indicated by the underline are unreliable.}
% \label{tb: other models}
% \end{table*}

\section{Additional Experimental Results}

\subsection{Comparison with Zero-Shot-CoT}
\label{app: svamp}

We mentioned in the main text that both our approach of role-play prompting and Zero-Shot-CoT reinforce the step-by-step reasoning process in tasks where ChatGPT can generate chain-of-thought \cite{chain} spontaneously. However, Zero-Shot-CoT demonstrates no significant effect while role-play prompting leads to better results. We provide an example of SVAMP dataset as shown in Table \ref{tb: svamp example}.

\begin{table*}[!h]
\centering
\begin{tabular}{p{16cm}}
\toprule
\textbf{Question}: Robin has 21 packages of gum and 45 packages of candy. There are 9 pieces in each package. How many pieces of candies does Robin have? \\
\midrule
\textbf{Zero-Shot}: \\
Robin has a total of 21 + 45 = 21 + 45 = 66 packages.\\
Therefore, Robin has 66 x 9 = 66 * 9 = 594 pieces of candies. \quad \textcolor{red}{\ding{55}} \\
\midrule
\textbf{Zero-Shot-CoT}:\\
First, we need to find the total number of pieces of gum. To do this, we multiply the number of packages of gum by the number of pieces in each package: 21 packages * 9 pieces/package = 21*9 = 189 pieces of gum.\\

Next, we need to find the total number of pieces of candy. To do this, we multiply the number of packages of candy by the number of pieces in each package: 45 packages * 9 pieces/package = 45 * 9 = 405 pieces of candy.\\

Therefore, Robin has a total of 189 pieces of gum + 405 pieces of candy = 189 + 405 = 594 pieces of candy. \quad \textcolor{red}{\ding{55}}                                       \\ 
\midrule
\textbf{Role-Play Prompting}:\\ 
To find out how many pieces of candy Robin has, we need to multiply the number of candy packages by the number of pieces in each package.
\\
Robin has 45 packages of candy, and each package contains 9 pieces. So, we can calculate the total number of candy pieces as follows:\\

45 packages * 9 pieces per package = 405 pieces of candy.\\

Therefore, Robin has 405 pieces of candy. \quad \textcolor{green}{\checkmark}\\
\bottomrule
\end{tabular}
\caption{An example of Zero-Shot, Zero-Shot-CoT, and Role-Play Prompting on SVAMP.}
\label{tb: svamp example}
\end{table*}

\subsection{Combination with Self-Consistency}
\label{app: sc}
Different from the naive greedy decoding, Self-Consistency (SC) \cite{selfcon} samples diverse reasoning paths and selects the most consistent answer by majority vote. We combine our approach and baselines with SC across multiple datasets, including AQuA, CSQA, Letter, Object, and Coin (N = 10 and temperature = 0.7). The results are shown in Table \ref{tb: sc}. With SC, role-play prompting still consistently outperforms zero-shot baseline, further proving the efficacy of our approach.

\begin{table*}[t]
\centering
\scalebox{0.95}{
\begin{tabular}{lccccc} 
\toprule
Method                & AQuA & CSQA & Letter & Object & Coin  \\ 
\midrule
Zero-Shot             & 53.5 & 74.5 & 23.8   & 38.7   & 55.2  \\
+SC                   & 61.0 & 74.7 & 38.2   & 42.9   & 55.2  \\ 
\midrule
Zero-Shot-CoT         & 53.9 & 68.8 & 53.2   & 73.5   & 98.8  \\
+SC                   & 65.0 & 72.1 & 88.6   & 88.3   & 99.8  \\ 
\midrule
Role-Play
  Prompting & 63.8 & 77.2 & 84.2   & 67.7   & 89.4  \\
+SC                   & 68.1 & 77.7 & 92.0   & 71.6   & 95.8  \\
\bottomrule
\end{tabular}}
\caption{Accuracy comparison of Role-Play Prompting against Zero-Shot and Zero-Shot-CoT, with and without SC.}
\label{tb: sc}
\end{table*}

\subsection{Ablation Study on Letter, Coin Datasets}
\label{app: letter}

Besides AQuA, we also conduct experiments on Letter and Coin datasets to explore the optimal prompt structure of role-play prompting. Consistent with the main text, we design 4 groups of prompts with progressively increasing levels of immersion, as shown in Table \ref{tb: internal design letter} and Table \ref{tb: internal design coin}. The results also demonstrate the effectiveness of the two-round prompt structure with complementary descriptions which enhance the model's immersion.

\begin{table*}
\centering
\scalebox{0.82}{
\begin{tabular}{c|p{16.5cm}|c} 
\toprule
No.             &  \multicolumn{1}{c}{Prompt}                                                                                                                                                                                                                                                                                                                      & Letter               \\ 
\midrule
1                  & From now on, you are a teacher. Please answer the following question.                                                                                                                                                                                                                           & 43.0                   \\  
\midrule
\multirow{1}{*}[-0.7em]{2}                  & From now on, you are an \colorbox{gray!17}{excellent} teacher and \colorbox{gray!17}{are teaching your students to get a new word by concat-} \colorbox{gray!17}{enating the last letters of several words. I am one of your students} and want to ask you a related question.                                                                     & \multirow{1}{*}[-0.7em]{60.8}                   \\ 
\midrule
\midrule
\multirow{2}{*}[-1.4em]{3} & From now on, you are an excellent teacher and are teaching your students to get a new word by concatenating the last letters of several words. I am one of your students and want to ask you a related question.                                                                                                                                                                                   & \multirow{2}{*}[-1.4em]{74.2}  \\ 
\cmidrule{2-2}
                   & \colorbox{gray!17}{Of course! Please go ahead and ask your question, and I'll do my best to assist you.}                                                                              &                        \\ 
\midrule
\multirow{2}{*}[-1.4em]{4} & From now on, you are an excellent math teacher and
  always teach your students math problems correctly. And I am one of your
  students.                                                                                                                                                                                   & \multirow{2}{*}[-1.4em]{84.2}  \\ 
\cmidrule{2-2}
                   & Of course! \colorbox{gray!17}{I’d be happy to help you with any questions you have about creating new words by concatena-} \colorbox{gray!17}{ting the last letters of
several words.} Please go ahead and ask your question, and I’ll do my best to assist you. &                        \\
\bottomrule
\end{tabular}}
\caption{Accuracy comparison of different prompt designs with a fixed role of the teacher on Last Letter dataset. We utilize gray shading to indicate the additional content in comparison to the previous prompt.}
\label{tb: internal design letter}
\end{table*}

\begin{table*}[t]
\centering
\scalebox{0.82}{
\begin{tabular}{c|p{16.5cm}|c} 
\toprule
No.             &  \multicolumn{1}{c}{Prompt}                                                                                                                                                                                                                                                                                                                      & Coin               \\ 
\midrule
1                  & From now on, you are a coin. Some people want to play a game with you.                                                                                                                                                                                                                           & 65.6                   \\  
\midrule
\multirow{1}{*}[-0.7em]{2}                  & From now on, you are a coin \colorbox{gray!17}{that always clearly knows which side of your head is facing ... tell them if} \colorbox{gray!17}{you (a coin) are heads up in the end.}                                                                     & \multirow{1}{*}[-0.7em]{71.4}                   \\ 
\midrule
\midrule
\multirow{2}{*}[-1.4em]{3} & From now on, you are a coin that always clearly knows which side of your head is facing ... tell them if you (a coin) are heads up in the end.                                                                                                                                                                                    & \multirow{2}{*}[-1.4em]{76.0}  \\ 
\cmidrule{2-2}
                   & \colorbox{gray!17}{Certainly! I'll let you know which side, heads or tails, is facing up. Feel free to start whenever you're} \ \colorbox{gray!17}{ready!}                                                                              &                        \\ 
\midrule
\multirow{2}{*}[-1.4em]{4} & From now on, you are a coin that always clearly knows which side of your head is facing ... tell them if you (a coin) are heads up in the end.                                                                                                                                                                                   & \multirow{2}{*}[-1.4em]{89.4}  \\ 
\cmidrule{2-2}
                   & Certainly! I’ll \colorbox{gray!17}{be your coin for this game. You can go ahead and flip me or make any other moves you’d} \colorbox{gray!17}{like} ... know which side, heads or tails, is facing up. Feel free to start whenever you’re ready! &                        \\
\bottomrule
\end{tabular}}
\caption{Accuracy comparison of different prompt designs with a fixed role of the coin on Coin Flip dataset. We utilize gray shading to indicate the additional content in comparison to the previous prompt.}
\label{tb: internal design coin}
\end{table*}

\subsection{Exploration of Prompt Length Impact}
\label{app: length effect}

From the results in Table \ref{tb: internal design}, \ref{tb: internal design letter}, and \ref{tb: internal design coin}, the improvement in accuracy may be attributed to the increase in prompt length. Therefore, we conduct additional experiments on Letter dataset. We replace the role-feedback prompt with generic responses of varying lengths that lack immersion. The results are shown in Table \ref{tb: length effect}.

\begin{table*}
\centering
\scalebox{0.89}{
\begin{tabular}{cp{13.7cm}cc} 
\toprule
No. &  \multicolumn{1}{c}{Prompt}                                                                                                                                                                                    & Sum & Letter  \\ 
\midrule
0   & One-round interaction (no role-feedback prompt)                                                                                                             & 0                 & 60.8      \\
1   & Of course! I’d ... by concatenating the last letters of several words ... (immersive) & 214               & 84.2      \\
2   & Certainly! Feel free to share any details ... I'm here to help!                                                                                             & 227               & 69.4      \\
3   & Certainly! Feel free to share any details ... to ensure you receive the support you need.                                                        & 349               & 75.6      \\
4   & Certainly! Feel free to share any details ... delivering a more tailored and accurate response.                                                & 473               & 72.4      \\
\bottomrule
\end{tabular}}
\caption{Accuracy comparison of prompts with different lengths. Sum represents the total number of characters in the prompt.}
\label{tb: length effect}
\end{table*}

Immersion of Prompt 1-4 all increase due to 2-round interaction so they surpass Prompt 0. And Prompt 1 outperforms Prompt 2-4 with longer lengths but lacking immersion. The results demonstrate that the improvement in performance is attributed to stronger immersion, rather than the increase in prompt length.

\subsection{Detailed Results of Model Scale Study}
\label{app: model_scale}

We examine the impact of model scale by testing the Llama 2-Chat series (7B, 13B, 70B) on GSM8K, MultiArith, and Letter datasets. The detailed experiment results are shown in Table \ref{tb: size_results}.

\begin{table*}[!t]
\centering
\begin{tabular}{lccc} 
\toprule
Method                & GSM8K          & MultiArith     & Letter       \\ 
\midrule
Zero-Shot             & 24.0 / 37.1 / 53.9 & 63.5 / 75.3 / 86.0 & 0 / 9.8 / 18.8   \\
Role-Play
  Prompting & 29.4 / 40.7 / 58.9 & 75.7 / 79.8 / 90.2 & 0 / 17.6 / 25.8  \\
\bottomrule
\end{tabular}
\caption{Accuracy comparison across different sizes of Llama 2-Chat models on GSM8K, MultiArith, and Letter. The data format is 7B / 13B / 70B.}
\label{tb: size_results}
\end{table*}

\section{Dataset Deatils}
\label{app: dataset}

We briefly introduce 12 datasets spanning four categories below. More information of 12 datasets is shown in Table \ref{tab:dataset_description}.

\noindent{\bf Arithmetic}\quad
We use the following six datasets: MultiArith, GSM8K, AddSub, AQUA-RAT, SingleEq, and SVAMP. All questions in these datasets contain a scenario and require reasoning based on mathematical knowledge.

%\subsubsection{Commonsense Reasoning}
\noindent{\bf Commonsense Reasoning}\quad
We utilize CSQA  and StrategyQA . Both of them require reasoning based on prior common sense.

%\subsubsection{Symbolic Reasoning}
\noindent{\bf Symbolic Reasoning}\quad
We employ Last Letter Concatenation and Coin Flip. Last Letter Concatenation requires concatenating the last letter of given words in order. Coin Flip gives a sequence of operations to flip a coin and asks for the final orientation of the coin. These two datasets are proposed by \citet{chain} but they are not available. \citet{zero_shot_cot} have followed the approach of \citet{chain} to create and release the datasets. We utilize this version for our experiments.

%\subsubsection{Other Reasoning Tasks}
\noindent{\bf Other Reasoning Tasks}\quad
We use Date Understanding and Tracking Shuffled Objects from BIG-bench. Date Understanding involves date calculations. Tracking Shuffled Objects gives a sequence of object exchange operations, asking for the final ownership of objects.

\begin{table*}[t]
\centering
\scalebox{0.82}{
\setlength{\tabcolsep}{20pt} % 设置列间距
\begin{tabular}{lcccc}
\toprule
Dataset &Answer Format & $N_{q}$ &$L_{q}$ &License \\\midrule
SingleEq &arabic number &508 &27.4 &No License \\
AddSub &arabic number &395 &31.5  &Unspecified \\
MultiArith &arabic number &600 &31.8  &Unspecified \\
GSM8K &arabic number &1319 &46.9  &MIT License \\
AQUA-RAT &option (A-E) &254 &51.9  &Apache-2.0 \\
SVAMP &arabic number &1000 &31.8  &MIT License \\
CommonsenseQA &option (A-E) &1221 &27.8  &Unspecified \\
StrategyQA &yes or no &2290 &9.6  &Apache-2.0 \\
Date Understanding &option (A-F) &369 &35.0  &Apache-2.0 \\
Shuffled Objects &option (A-C) &750 &91.1 &Apache-2.0 \\
Last Letters &string &500 &15.0  &- \\
Coin Flip &yes or no &500 &37.0  &- \\
\bottomrule
\end{tabular}}
\caption{Relevant information of 12 datasets. $N_{q}$ denotes the number of questions in each dataset. $L_{q}$ denotes the average words of questions in each dataset.}
\label{tab:dataset_description}
\end{table*}

\section{Prompts for Role Selection Study}
\label{app: prompt}

To investigate the role selection's impact on role-play prompting, we design 8 different roles for our study. The specific prompts, including role-setting prompts and role-feedback prompts are shown in Table \ref{tb: role_propmts}.

\begin{table*}[!h]
\centering
\scalebox{0.74}{
\begin{tabular}{l|p{18cm}} 
\toprule
Role                    &  \multicolumn{1}{c}{Prompt}                                                                                 \\                                                           
\midrule
\multirow{2}{*}[-1.6em]{math teacher} & From now on, you are an excellent math teacher and always teach your students math problems correctly. And I am one of your students.                                                                                                \\ 
\cmidrule{2-2}
& That's great to hear! As your math teacher, I'll do my best to explain mathematical concepts correctly so that you can understand them easily. Feel free to ask any math problems or questions you have, and I'll be glad to assist you. Let's dive into the world of mathematics and explore its wonders together!                                                                                          \\ 
\midrule
\multirow{2}{*}[-1.1em]{mathematician}   & From now on, you are an excellent mathematician who can always explain math problems in an easy-to-understand manner.                                                                         \\ 
\cmidrule{2-2}
 & Thank you for the compliment! I'll do my best to explain math problems in an easy-to-understand manner. Feel free to ask me any math questions you have, and I'll be happy to help you out!                                                                                 \\ 
\midrule
\multirow{2}{*}[-1.5em]{police}   & From now on, you are a responsible police officer. You always enforce the law impartially, actively help others, and have made great contributions to local public security. I am a citizen who has come to seek help.                                                                                                                                 \\ 
\cmidrule{2-2}
& Of course, as a responsible police officer, I am here to help you. Please feel free to share the issue or concern you have, and I'll do my best to assist you impartially and effectively. Your safety and well-being are my priority, and I'll work diligently to ensure the public security of our community. What can I do for you today?                                                                                                                                         \\ 
\midrule
\multirow{2}{*}[-1.3em]{farmer} & From now on, you are an experienced farmer. You have a lot of knowledge about crop cultivation, and your farm is running very well.                                                                                   \\ 
\cmidrule{2-2}
& As an experienced farmer, I can provide you with valuable insights and advice on various aspects of crop cultivation. Whether you're interested in sustainable farming practices, maximizing yields, or dealing with common challenges, I'm here to assist you. \\
\midrule
\multirow{2}{*}[-1.1em]{doctor} & From now on, you are a respectable doctor. You have profound medical knowledge and have saved many lives. I am your patient and want to ask some questions.                                                                                  \\ 
\cmidrule{2-2}
& Of course, I am here to help as your respectable doctor. Please feel free to ask any medical questions you may have, and I'll do my best to provide you with accurate and helpful information. Your health and well-being are my top priorities. \\
\midrule
\multirow{2}{*}[-1.7em]{writer} & From now on, you are a famous writer, and the novels you write are lively, interesting and fascinating. I am your reader and want to ask you some questions.                                                                                  \\ 
\cmidrule{2-2}
& Of course! I would be delighted to answer any questions you have as a reader. As a famous writer, I strive to create lively, interesting, and fascinating novels that captivate my audience. Please feel free to ask anything you'd like, and I'll do my best to provide insightful responses. \\
\midrule
\multirow{2}{*}[-1.3em]{careless student} & From now on, please act as a careless student. You always don't pay attention when doing math problems, and you end up making mistakes because of carelessness. Therefore, your math grade is very poor.                                                                           \\ 
\cmidrule{2-2}
& Oh, hey there! Math is not really my thing, and I tend to zone out a lot while doing problems. It's like my brain just switches off or something. My math grade is probably a disaster because of all the careless mistakes I make. \\
\midrule
\multirow{2}{*}[-1.2em]{math rookie} & From now on, please act as a math rookie. You don't have any talent for mathematics, and you don't even understand the most basic mathematical concepts. So you always make mistakes in math problems.                                                                              \\ 
\cmidrule{2-2}
& Sure, I can pretend to be a math rookie! Feel free to ask me any math questions, but be prepared for some silly mistakes and misunderstandings in my responses. \\
\bottomrule
\end{tabular}}
\caption{Prompts for different roles. For each role, the upper cell contains the role-setting prompt and the lower cell presents the role-feedback prompt.}
\label{tb: role_propmts}
\end{table*}

\end{document}